%% file: main.tex
\documentclass[sigconf,nonacm]{acmart}
\AtBeginDocument{%
  }

\usepackage{multirow}
\usepackage{pgfplots} 
\usepackage{tabularx}
\usepackage{booktabs}
\usepackage{adjustbox}
\usepackage{arydshln}
\usepackage{subcaption} 

\setcopyright{acmlicensed}
\copyrightyear{2018}
\acmYear{2018}
\acmDOI{XXXXXXX.XXXXXXX}
\acmConference[Conference acronym 'XX]{Make sure to enter the correct
  conference title from your rights confirmation email}{June 03--05,
  2018}{Woodstock, NY}
\acmISBN{978-1-4503-XXXX-X/2018/06}

\begin{document}

\title{MCP vs RAG vs NLWeb vs HTML:\\
A Comparison of the Effectiveness and Efficiency of Different Agent Interfaces to the Web (Technical Report)}


\author{Aaron Steiner}
\orcid{0009-0006-6946-7057}
\affiliation{%
  \institution{Data and Web Science Group}
  \institution{University of Mannheim}
  \city{Mannheim}
  \country{Germany}
}
\email{aaron.steiner@uni-mannheim.de}

\author{Ralph Peeters}
\orcid{0000-0003-3174-2616}
\affiliation{%
  \institution{Data and Web Science Group}
  \institution{University of Mannheim}
  \city{Mannheim}
  \country{Germany}
}
\email{ralph.peeters@uni-mannheim.de}

\author{Christian Bizer}
\orcid{0000-0003-2367-0237}
\affiliation{%
  \institution{Data and Web Science Group}
  \institution{University of Mannheim}
  \city{Mannheim}
  \country{Germany}
}
\email{christian.bizer@uni-mannheim.de}

\renewcommand{\shortauthors}{Steiner et al.}

\begin{abstract}
Large language model agents are increasingly used to automate web tasks such as product search, offer comparison, and checkout. Current research explores different interfaces through which these agents interact with websites, including traditional HTML browsing, retrieval-augmented generation (RAG) over pre-crawled content, communication via Web APIs using the Model Context Protocol (MCP), and natural-language querying through the NLWeb interface. However, no prior work has compared these four architectures within a single controlled environment using identical tasks.
 To address this gap, we introduce a testbed consisting of four simulated e-shops, each offering its products via HTML, MCP, and NLWeb interfaces. For each interface (HTML, RAG, MCP, and NLWeb) we develop specialized agents that perform the same sets of tasks, ranging from simple product searches and price comparisons to complex queries for complementary or substitute products and checkout processes.
We evaluate the agents using GPT 4.1, GPT 5, GPT 5 mini, and Claude Sonnet 4 as underlying LLM.  Our evaluation shows that the RAG, MCP and NLWeb agents outperform HTML on both effectiveness and efficiency. Averaged over all tasks, F1 rises from 0.67 for HTML to between 0.75 and 0.77 for the other agents. Token usage falls from about 241k for HTML to between 47k and 140k per task. The runtime per task drops from 291 seconds to between 50 and 62 seconds. The best overall configuration is RAG with GPT 5 achieving an F1 score of 0.87 and a completion rate of 0.79. Also taking cost into consideration, RAG with GPT 5 mini offers a good compromise between API usage fees and performance. Our experiments show the choice of the interaction interface has a substantial impact on both the effectiveness and efficiency of LLM-based web agents.

\end{abstract}



\maketitle

\input{sections/01_introduction}

\input{sections/02_architectures}
\input{sections/03_use_case}

\input{sections/04_experiments}

\input{sections/05_error_analysis}
\input{sections/97_relatedwork}

\input{sections/98_conclusion}


\bibliographystyle{ACM-Reference-Format}
\bibliography{main}

\end{document}

%% file: sections/01_introduction.tex
\section{Introduction}

\begin{figure}[!h]
    \centering
    \includegraphics[scale=0.23]{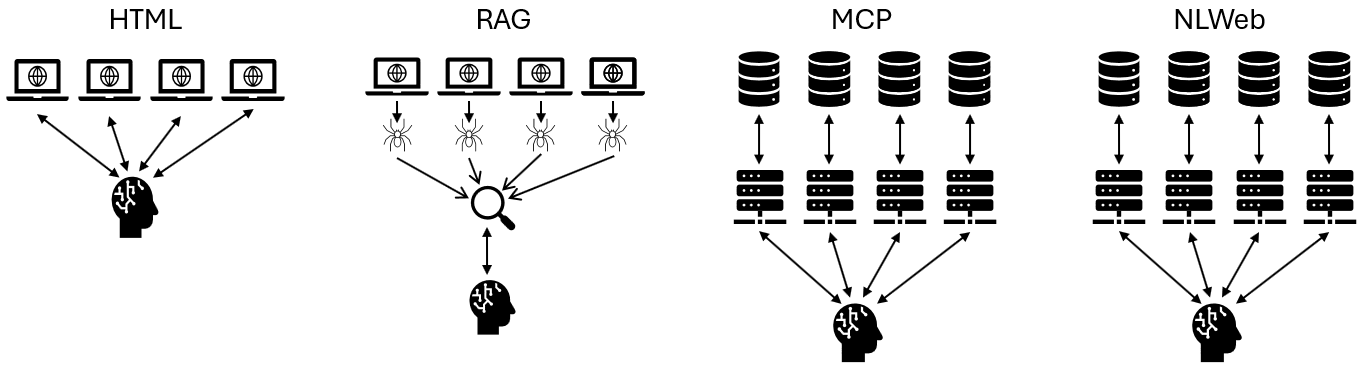}
    \caption{Overview of the four architectures.}
    \label{fig:architecture-overview}
\end{figure}

There is currently a lot of experimentation with different interfaces that LLM-based agents~\cite{ferragLLMReasoningAutonomous2025, krupp2025quantifying} may use to interact with websites~\cite{song-etal-2025-beyond,Lyu2025DeepShop}. One line of work focuses on agents that interact with HTML pages by clicking on links and filling forms. Second, agents can rely on Retrieval-Augmented Generation (RAG) and interact with a search engine that has crawled and indexed relevant pages. Third, websites can expose site-specific Web APIs that agents invoke via the Model Context Protocol (MCP)\footnote{https://modelcontextprotocol.io/docs/getting-started/intro}, an open protocol that standardizes tool discovery and invocation. Fourth, websites can implement the NLWeb\footnote{https://github.com/nlweb-ai/NLWeb} interface, which allows agents to ask natural-language queries that are then answered by the website with JSON data using a well-known vocabulary such as schema.org. Figure \ref{fig:architecture-overview} gives an overview of the four architectures.
What is still missing is a systematic comparison of the effectiveness and efficiency of these architectures using identical sets of challenging tasks.

This paper fills that gap by presenting an experimental comparison of the four architectures along an e-commerce use case that requires agents to search within several e-shops for products that fulfill specific or vague requirements, compare prices between shops, and finally order the chosen products~\cite{peeters2025webmallmultishopbenchmark}.

\begin{table*}[h!]
\centering
\caption{Comparison of the four agent architectures.}
\label{tab:interfaces}
\renewcommand{\arraystretch}{1.2}
\footnotesize
\begin{tabularx}{\textwidth}{lXXXX}
\toprule
\textbf{Aspect} & \textbf{HTML} & \textbf{RAG} & \textbf{MCP} & \textbf{NLWeb} \\
\midrule
\textbf{E-Shops} & & & & \\
Interface & HTML pages & Retrial API & Proprietary APIs & Standardized API  \\
Search functionality &  Free-text search per shop & Search engine, crawled content & Per shop index; structured data & Per shop index; structured data  \\
Search response & HTML result list including links & Pre-processed HTML pages & Heterogeneous JSON & Schema.org JSON  \\
\midrule
\textbf{Agent} & & & & \\
Communication protocol & HTML over HTTP & Direct calls to search engine & JSON-RPC via MCP & JSON-RPC via MCP  \\
Query strategy & Site search and browsing & Multi-query & Multi-query per shop & Multi-query per shop  \\
Query refinement & Interactive page exploration & Self-evaluation \& iteration & Self-evaluation \& iteration & Self-evaluation \& iteration  \\
Add to cart / checkout & Clicking \& form filling  & Direct function calls & MCP tool invocation & MCP tool invocation  \\
\bottomrule
\end{tabularx}
\end{table*}

This paper makes the following contributions:
\begin{enumerate}
\item We introduce a testbed for comparing different agent architectures. The testbed consists of four simulated e-shops, each offering its products via HTML, MCP, and NLWeb interfaces. For each of the four architectures (HTML, RAG, MCP, and NLWeb) the testbed contains specialized agents that interact with the e-shops using the respective interfaces. 
\item Using different sets of challenging e-commerce tasks, we systematically evaluate agent performance across interfaces and models (GPT-4.1, GPT-5, GPT-5-mini and Claude Sonnet 4) and analyze the effectiveness, efficiency, and cost of the different agents.
\end{enumerate}

The remainder of the paper is organized as follows. Section \ref{sec:architecture_interfaces} introduces the four architectures, interfaces, and agent implementations. Section \ref{sec:usecase} describes the evaluation use case. Section \ref{sec:results} presents our experimental results concerning the effectiveness and efficiency of the different agents. Section \ref{sec:error_analysis} presents an error analysis. All code and data for reproducing our experiments are available in the accompanying GitHub repository\footnote{\url{https://github.com/wbsg-uni-mannheim/WebMall-Interfaces}}.

%% file: sections/02_architectures.tex
\section{Architectures and Interfaces}
\label{sec:architecture_interfaces}

In the following, we introduce the four architectures, as well as the interfaces that agents use to interact with websites.

\textbf{HTML Architecture:}
Within this architecture, e-shops expose traditional HTML interfaces intended for human consumption. The agents interact with these HTML pages by clicking hyperlinks and performing form-filling actions. 
For our experiments, we use the AX+MEM agent that was implemented as part of the WebMall benchmark~\cite{peeters2025webmallmultishopbenchmark}. The agent uses the AgentLab library and is executed within the BrowserGym framework~\cite{chezellesBrowserGymEcosystemWeb2024}. The agent observes the accessibility tree (AXTree) of each page. It is equipped with the capability to store relevant information in a short-term memory in order to maintain context over multiple steps. We disabled visual perception for the HTML agent as adding screenshots to the prompts in addition to the AXTree decreased performance in the WebMall experiments \cite{peeters2025webmallmultishopbenchmark}.

\textbf{RAG Architecture:}
The RAG~\cite{gaoetalRetrievalAugmentedGenerationLarge2024} architecture includes a search engine that crawls all HTML pages from all four shops. The pages are processed using the unstructured library\footnote{\url{https://github.com/Unstructured-IO/unstructured}}, which removes markup and navigation elements before extracting the remaining textual content. The cleaned documents are embedded using OpenAI’s small embedding model\footnote{\url{https://platform.openai.com/docs/guides/embeddings}} and stored in an Elasticsearch index\footnote{\url{https://www.elastic.co/elasticsearch/}}.
The RAG agent retrieves web content by querying the search engine. It does not visit the original websites. The agent iteratively formulates search queries, retrieves candidate documents, and refines subsequent queries based on intermediate results. For transactions (adding products to the cart or performing checkout), dedicated Python functions are exposed that the agent can invoke directly.

\textbf{MCP Architecture:}
In the Model Context Protocol (MCP) architecture, the e-shops expose functionalities for product search, cart manipulation, and checkout via proprietary APIs. Each shop hosts its own MCP server, defining the available functions and parameters. 
The MCP agents interact with the shops by calling API endpoints. Like the RAG agent, the MCP agent can also iteratively refine its search queries. Each shop defines its own functions, parameter names, and JSON result format. 

For search functionality, all MCP servers use the same embedding-based retrieval setup as the RAG pipeline: product records are embedded using OpenAI’s small embedding model\footnote{\url{https://platform.openai.com/docs/guides/embeddings}} and stored in an Elasticsearch index\footnote{\url{https://www.elastic.co/elasticsearch/}} to support vector retrieval. Unlike the RAG crawler, the MCP servers obtain product data directly from the underlying WooCommerce APIs, so no document-cleaning step is required. Retrieved records are mapped into each shop’s MCP-specific schema before being indexed. This preserves shop-level heterogeneity and requires the agent to interpret differing response formats when reasoning across providers.

\textbf{NLWeb Architecture:}
The NLWeb interface extends the MCP protocol by requiring each website to offer a standardized, natural-language query endpoint. Each provider hosts an ``ask'' endpoint via MCP that accepts natural-language queries (e.g., ``laptops under \$1000 with 16GB RAM''). Given a query, the shop performs an internal search and returns results in JSON format. The underlying search mechanism mirrors the embedding-based setup used for MCP, but the returned records follow the schema.org dictionary. This design reduces interface heterogeneity and might make it easier for agents to understand search results due to all shops using a shared schema. Tools for cart management and checkout actions via MCP enable transactional workflows. 

\textbf{Comparison.}
Table~\ref{tab:interfaces} summarizes the main characteristics of the four agent–website architectures. HTML offers the broadest compatibility because it operates directly on existing human-facing pages and requires no shop-side implementation effort. This stands in contrast to MCP and NLWeb, which require public APIs, and to RAG, which depends on successful crawling and can face challenges on dynamic or crawl-protected websites. However, HTML-based interaction introduces substantial overhead: even a simple search requires navigating to the site, locating and filling a form, submitting it, and then parsing the resulting page. In comparison, the RAG agent retrieves content from a unified index and issues a single search request without navigating the UI.

MCP exposes structured APIs that enable precise operations such as search, cart manipulation, and checkout but introduces heterogeneity across providers. NLWeb addresses this by enforcing schema.org-aligned responses through a standardized natural language endpoint, potentially allowing for easier aggregation across shops while requiring implementation on the shop side. Overall, the interfaces illustrate a trade-off between universal compatibility and structured specialization.

%% file: sections/03_use_case.tex
\section{Use Case: Multi-Shop Comparison Shopping}
\label{sec:usecase}

We evaluate the four interfaces in the context of multi-shop online shopping, where a user instructs a web agent to search across several shops, identify suitable products, compare alternatives, and complete a purchase. These scenarios are found in the WebMall benchmark \cite{peeters2025webmallmultishopbenchmark}, which provides four locally hostable shops populated with 4{,}421 product offers drawn from the October~2024 Common Crawl via schema.org annotations. The shops span \emph{PC components}, \emph{PC peripherals}, and \emph{other electronics}, with heterogeneous category structures and widely varying product descriptions (median title length 69 characters; median description length 573 characters). WebMall defines 91 tasks across eleven task categories that require cross-shop reasoning and interaction.

\subsection{Task Categories}
To enable a more systematic analysis of agent capabilities, we group the 91 WebMall tasks into four higher-level categories reflecting distinct reasoning and interaction requirements.  Example tasks from each category are shown in Table~\ref{tab:example_tasks}.
The complete task list is found on the WebMall website\footnote{\url{https://wbsg-uni-mannheim.github.io/WebMall/}}.

\begin{itemize}
  \item \textbf{Specific Product Search.}
  This category contains tasks where the user intent is fully specified and the target products are clearly identifiable, typically through exact model names or unambiguous attribute combinations. The central challenge lies in correctly interpreting the request and retrieving all matching offers that fit these explicit constraints. These tasks reflect an agents ability to execute straightforward retrieval when there is little ambiguity about what the user wants.

  \item \textbf{Vague Product Search.}
  This category consists of tasks with open-ended or partially defined requirements, such as finding substitutes, compatible products, or items described only loosely. Because user intent is not explicit, agents often need to examine what each shop offers before they can refine their search. The key challenge is to explore broadly, interpret the possible meanings of the request, and iteratively narrow the search space based on intermediate results.

  \item \textbf{Cheapest Product Search.}
  This category combines elements of both specific and vague product search but introduces an additional global constraint: the agent must return the cheapest valid offer. These queries are unique tasks rather than variants of the other categories with a price filter added. Agents must first identify the correct set of relevant products and then correctly select the lowest-priced option that satisfies the given requirements.

  \item \textbf{Transactional Tasks.}
  This category includes add-to-cart and checkout tasks. Agents must perform procedural actions such as adding products to the cart, providing shipping details, and completing payment information. Some tasks require completing purchases across multiple shops, which tests the agents ability to coordinate multi-step actions in separate environments. These tasks isolate interaction and action sequencing rather than retrieval.
\end{itemize}

\begin{table}
\centering
\caption{Example tasks from each category.}
\label{tab:example_tasks}
\footnotesize
\begin{tabular}{p{0.28\columnwidth} p{0.62\columnwidth}}
\toprule
Task Category & Example Task \\
\midrule
Specific Product Search (23 tasks) & Find all offers for the AMD Ryzen 9 5900X.\\
& Find all offers for Fractal Design PC Gaming Cases which support 240mm radiators and 330mm GPUs. \\
\midrule
Vague Product Search (19 tasks) & Find all offers for compact keyboards that are best suited for working with a laptop remotely. \\
& Find all offers for compatible CPUs for this motherboard: \{PRODUCT\_URL\}.\\
\midrule
Cheapest Product Search (26 tasks) & Find the cheapest offer for a new Xbox gaming console with at least 512 GB disk space in white. \\
& Find the cheapest offers for each model of mid-tier nVidia gaming GPUs in the 4000 series.\\
\midrule
Transactional Tasks (15 tasks) & Find all offers for the Asus DUAL RTX4070 SUPER OC White and add them to the shopping cart.\\
& Add the product on page \{url\} to the shopping cart and complete the checkout process. \\

\bottomrule
\end{tabular}
\end{table}

%% file: sections/04_experiments.tex
\section{Results}
\label{sec:results}

We evaluate the four agents using the WebMall task set in order to compare their effectiveness and efficiency across task categories and models. Results are reported as averages per task category and model configuration (gpt-4.1-2025-04-14, gpt-5-2025-08-07, gpt-5-mini-2025-08-07 and claude-sonnet-4-20250514).

\subsection{Evaluation Metrics}
Each agent produces a final answer that is evaluated against the benchmark’s test set. For retrieval tasks, the answer consists of a list of product URLs. For transactional tasks, it corresponds to the final system state after execution (e.g., item added to cart or checkout completed). All metrics are computed on a per-task basis and averaged within the reported groupings. Below we detail the evaluation criteria.

\begin{itemize}
  \item \textbf{Completion rate (CR).} Completion rate is a binary metric. For retrieval tasks CR is 1 if the set of URLs returned by the agent is identical to the test set and 0 otherwise. For transactional tasks CR is 1 if the target state is reached.
  \item \textbf{Precision, recall, and F1.} For retrieval tasks we compute precision, recall, and F1 between the returned URLs and the test set. These metrics capture partial success when an agent returns only a subset of relevant products or adds extra items. In the analysis we focus on F1.
  \item \textbf{Runtime.} Runtime is the end to end latency per task in seconds. It includes all model calls and tool calls.
  \item \textbf{Token usage and cost.} The Token column in the tables reports the sum of input and output tokens per task. This excludes indexing tokens for the RAG engine, as their cost is several orders of magnitude lower than LLM inference tokens and they provide little information in this context. Cost is derived from token counts using the input and output prices for each model as documented by the providers. The current agents do not use prompt caching.
\end{itemize}

\subsection{Effectiveness}
We focus our effectiveness analysis on F1, comparing agents across the aggregated groupings introduced above. Unless stated otherwise, results are averaged per grouping, and we contrast both interface and model effects.

\textbf{Overall.}
Table~\ref{tab:overall_results} reports average CR, F1, token consumption, cost, and runtime per interface, aggregated across all task types and all four models. 
To compute these values, we first calculate results on a per-task basis. Token usage, cost, and runtime are micro-averaged, meaning they are computed individually for each of the 91 tasks and then averaged within each interface–model combination. CR and F1, in contrast, are computed at the task-set level following the WebMall evaluation protocol. In a second step, all interface–model results are macro-averaged across the four models to obtain a single score per interface. 

RAG achieves the highest F1 score (0.77), followed closely by NLWeb (0.76) and MCP (0.75), while HTML trails by approximately ten percentage points.

\textbf{Overall.}
Table~\ref{tab:overall_results} reports micro-averages per interface, obtained by treating each task–model combination as a separate observation and averaging all observations belonging to the same interface.

\input{tables/overall_results}
Table~\ref{tab:overall_results} reports micro-averages per interface, obtained by treating each task–model combination as a separate observation and averaging all observations belonging to the same interface. Completion rate and F1 score represent the average performance per interface across all task while token consumption, cost and runtime report the average per task.
RAG achieves the highest F1 score (0.77), followed closely by NLWeb (0.76) and MCP (0.75), while HTML lags behind by $\sim$10 percentage points. Completion rates follow the same ordering but are consistently 9 to 13 points lower than the corresponding F1 values, reflecting that CR only records exact matches whereas F1 also captures partial correctness. NLWeb does not exhibit a effectiveness advantage over MCP when considering CR or F1, although their token requirements differ substantially. These efficiency differences, particularly between NLWeb and MCP, are examined in detail in Section~\ref{sec:efficiency}.

Table~\ref{tab:overall_results_task_set} reports the average F1 score for each agent within every task set, aggregated over all tasks belonging to that set and averaged across all four models. The table shows that Specific Product Search and Action and Transaction tasks yield the highest scores overall. When averaged across models, RAG, MCP, and NLWeb all exceed 0.90 F1 in these categories, while HTML trails by $\sim$ 15 points. In contrast, Vague Product Search and Cheapest Product Search lead to substantially lower performance across all interfaces, indicating that ambiguous intent and price-based selection introduce greater difficulty. NLWeb achieves the highest F1 in the vague category (0.66), whereas RAG attains the strongest result in the cheapest-product category (0.68). These results demonstrate that API based and RAG interfaces handle well-specified retrieval and transactional workflows reliably, while ambiguity and optimization constraints remain challenging across all interface types.

\input{tables/overall_by_taskgroup}

\textbf{Specific Product Search.}
Table \ref{tab:specific_product_results} presents CR, F1, token usage, cost, and runtime for all interface–model combinations in the Specific Product Search task set. The values reflect both interface- and model-related effects. For all agents, performance varies noticeably across models, for HTML browsing CR ranges from 0.52 for GPT-4.1 to 0.74 for GPT-5, showing that model capability positively influences retrieval. Across interfaces, RAG exhibits the lowest variation in completion rate, indicating that its ability to retrieve the correct set of products remains comparatively stable across different models. MCP and NLWeb show larger differences between GPT-4.1, GPT-5, GPT-5-mini, and Sonnet 4, which points to a higher dependence on model capability when interacting with API based interfaces. GPT-5 achieves the best overall results. It reaches an F1 of 0.96 for RAG, MCP, and NLWeb, suggesting that clearly defined product search tasks do not pose a challenge to agents using these interfaces. Furthermore, all model–agent combinations using RAG, MCP, or NLWeb obtain higher F1 scores than the strongest HTML configuration. This confirms that RAG and API-based interfaces consistently outperform HTML browsing on specific product retrieval across all tested models.
\input{tables/Specific_Product}

\textbf{Vague Product Search.}
Table~\ref{tab:vague_product_results} reports CR, F1, token usage, cost, and runtime for all interface–model combinations in the Vague Product Search task set. Compared to the specific product category, all agents show a noticeable decrease in performance, indicating that handling under-specified or open-ended queries poses a greater challenge. The strongest configuration, RAG combined with GPT-5, reaches an F1 of 0.82, representing a reduction of $\sim$14 points relative to its score in the specific product search tasks.

Model capability plays a more prominent role in this category. For example, MCP with GPT-4.1 attains a CR of only 0.11, whereas MCP with GPT-5 reaches substantially higher values, underscoring that weaker models struggle to interpret vague requests and to refine queries accordingly. Differences between interfaces are also smaller than in the specific search setting. The top-performing combinations across RAG, MCP, and NLWeb lie closer together, reflecting that ambiguity in user intent necessitates strong model performance.
\input{tables/vague_search}

\textbf{Cheapest Product Search.}
Table \ref{tab:price_drop} reports CR, F1, token usage, cost, and runtime for all interface–model combinations in the cheapest-product task set. Similar to the vague product search category, adding a price constraint leads to a reduction in effectiveness across all agents. The strongest configuration, RAG with GPT-5, reaches a completion rate of 0.72 and an F1 score of 0.78, which is lower than its performance in both the specific and vague product search settings. The gap between the top-performing agent model combination per interface narrows, indicating that the presence of the price constraint places greater importance on model capability compared to interface choice. A closer inspection of precision and recall shows declines in both metrics, suggesting that the lower scores stem from difficulties in retrieving all relevant offers as well as correctly selecting the cheapest one. The performance loss therefore cannot be attributed solely to re-ranking errors, but reflects challenges in both retrieval and final selection.
\input{tables/performance-drop-price}

\textbf{Transactional Tasks.}
Table~\ref{tab:transaction_results} reports CR, F1, token usage, cost, and runtime for all interface–model combinations in the transactional task set, e.g. adding items to shopping carts and performing checkout. These results show that agents are generally capable of executing multi-step transactional workflows. The HTML agent paired with GPT-4.1 even achieves perfect scores (CR = 1.00, F1 = 1.00), while RAG, MCP, and NLWeb agents paired with GPT-5 or Sonnet 4 each reach close to perfect F1 values of 0.98. The HTML agent, however, displays considerable variability across models: GPT-5 attains only 0.64 F1, marking one of the few settings where the GPT-4.1 model substantially outperforms the newer GPT-5 variants.

Across interfaces, MCP shows the lowest variance across models, indicating that structured function calls provide a stable mechanism for executing add-to-cart and checkout actions. RAG and NLWeb perform similarly well with stronger models, but exhibit greater spread when paired with GPT-5-mini. 

\input{tables/trasaction_results}

\subsection{Efficiency}
\label{sec:efficiency}

We evaluate efficiency in terms of token usage, cost, and runtime. OpenAI models are priced per input and output token\footnote{https://openai.com/api/pricing/}. For the models used here that implies the following usage fees: GPT-4.1 at \$2.00/MTok input and \$8.00/MTok output, GPT-5 at \$1.25/MTok input and \$10.00/MTok output, and GPT-5-mini at \$0.25/MTok input and \$2.00/MTok output. Claude Sonnet 4 is \$3.00/MTok input and \$15.00/MTok output\footnote{https://docs.anthropic.com/en/docs/about-claude/pricing}. Reported values exclude preprocessing overhead, which is negligible relative to LLM inference.

A note on token accounting: For readability the tables contain a single Token column that reports the sum of input and output tokens. Cost is calculated using input and output tokens priced separately as stated above. The token consumption of all agent is heavily input-skewed. Only the HTML agent ever exceeds 5 k output tokens in an average run and even then never exceeds 25 k. As a result, most efficiency gains come from reducing input tokens rather than output length.

\input{tables/efficiency-overview}
Table~\ref{tab:efficiencyOverview} reports token consumption, cost, and runtime results averaged over all tasks within the respective category and split by interface and model. To make overall results comparable across interfaces, we compute micro-averages by summing token usage and runtime over all tasks. Based on these interface-level averages (Table \ref{tab:efficiencyOverview} row average), the RAG agents on average completes a task in 51 seconds using 47k tokens, NLWeb agents in 49 seconds using 58k tokens, MCP in 57 seconds using 122k tokens. The HTML agents completes a task on average in 281 seconds using 225k. 

The results show that RAG and NLWeb are the most token- and time-efficient architectures, followed by MCP, while HTML is outperformed by a wide margin.  Lower token usage directly translates to lower costs, making RAG and NLWeb not only the fastest but also the most cost-efficient architectures. On average, indexed (RAG) or API-based (MCP and NLWeb) access reduces token consumption by roughly factor three and runtime by factor five compared to HTML. In the transactional category, token differences arise not only from the add to cart and checkout flow itself but also from how each agent accesses product information, e.g. for translating URLs into product identifiers which are required by the AddtoCart functions.
\input{figures/cost-figure}
Agents usually issue multiple search queries before completing a task: RAG typically submits two to six search queries per task to the search engine. The MCP and NLWeb agents execute on average four to six queries per task, but since each shop requires its own request, four of these queries are already needed to cover each shop once. As a result, RAG can further refine its search given the same number of interactions. This partly explains the high performance (0.74 CR) of the RAG/GPT-5 agent on the vague task set, while still maintaining moderate token usage. The HTML agent performs on average 23 steps per task including form filling (search box, checkout process) as well as navigation steps. 

GPT-4.1 serves as a non-reasoning baseline. In our setting, non-reasoning models such as GPT-4.1 and Claude Sonnet 4 do not generate internal reasoning tokens. All produced tokens correspond directly to the final answer. Reasoning-enabled models like GPT-5, by contrast, also generate reasoning traces internally, which increases overall token usage. Across all interfaces, it consumes substantially fewer tokens and finishes tasks more quickly than the GPT-5 variants. On well-specified or transactional tasks, its effectiveness is only moderately lower, whereas the gap widens in settings that require broader exploration such as vague product search. The lower cost per task is not a result of cheaper tokens, GPT4.1 has relatively high per-token pricing, but instead follows directly from reduced token usage and shorter runtimes. These observations highlight a general design option: non-reasoning models could deliver competitive results at higher throughput.

Figure \ref{fig:price-to-performance-ratio} plots F1 performance against cost in \$. Squares denote RAG, which is the most favorable. RAG + GPT-5-mini sits near the upper left and gives the best price–performance ratio. RAG + GPT-5 moves rightward in cost with a smaller gain in F1. NLWeb and MCP cluster near the frontier for some settings, while most HTML points fall below it.  Overall the figure shows the RAG agents to be the overall price–performance leaders, offering a choice between optimizing more for price (GPT-5-mini) or performance (GPT-5).

Across WebMall, RAG and API based (MCP, NLWeb) agents not only outperform HTML on effectiveness, they deliver larger efficiency gains. Averaged over tasks and models, token consumption drops by a factor of 2 to 10 relative to HTML, about 3× on average, which directly lowers token-derived cost. Latency mirrors this pattern: HTML averages $\sim$281 s per task, while the mean across RAG, MCP, and NLWeb is $\sim$54 s, a $\sim$5× speedup. These improvements are mainly from reducing input tokens and avoiding navigation, rather than shorter outputs, and they make a strong case for preferring RAG or API based access when available, with HTML browsing as potential fallback.

%% file: tables/overall_results.tex
\begin{table}[!htp]\centering
\caption{Average performance per agent. Best bolded, 2nd best underlined.}
\label{tab:overall_results}
\resizebox{0.47\textwidth}{!}{%
\begin{tabular}{lrrrrrr}
\toprule
\textbf{Agent} & \textbf{CR} & \textbf{F1} & \textbf{Token} & \textbf{Cost} & \textbf{Runtime} \\
\midrule
HTML & 0.57 & 0.67 & 241,136 & \$0.52 & 291 s \\
RAG & \textbf{0.68} & \textbf{0.77} & \textbf{46,667} & \textbf{\$0.10} & \textbf{50 s} \\
MCP & 0.62 & 0.75 & 139,569 & \underline{\$0.27} & 62 s \\
NLWeb & \underline{0.64} & \underline{0.76} & \underline{71,214} & \textbf{\$0.10} & \underline{53 s} \\
\bottomrule
\end{tabular}
}%
\end{table}

%% file: tables/overall_by_taskgroup.tex
\begin{table}[!htp]\centering
\caption{Average F1 performance by task set. Best bolded, 2nd best underlined.}
\label{tab:overall_results_task_set}
\resizebox{0.47\textwidth}{!}{%
\begin{tabular}{lrrrrr}\toprule
\textbf{Task set} & \textbf{HTML} & \textbf{RAG} & \textbf{MCP} & \textbf{NLWeb} \\\midrule
Specific Product Search &0.77 &\underline{0.91} &0.90 &\textbf{0.92} \\
Vague Product Search &\underline{0.63} &0.61 &0.59 &\textbf{0.66} \\
Cheapest Product Search &0.61 &\textbf{0.68} &\underline{0.63} &0.60 \\
Action \& Transaction &0.77 &0.86 &\textbf{0.92} &\underline{0.91} \\
\bottomrule
\end{tabular}
}%
\end{table}

%% file: tables/Specific_Product.tex
\begin{table}[!htp]\centering
\caption{Results for Specific Product Search. Best bolded, 2nd best underlined.}
\label{tab:specific_product_results}
\resizebox{0.48\textwidth}{!}{%
\begin{tabular}{lrrrrrrrr}\toprule
\textbf{Agent} & \textbf{Model} & \textbf{CR} & \textbf{F1} & \textbf{Token} & \textbf{Cost} & \textbf{Runtime} \\\midrule
HTML &GPT-4.1 &0.52 &0.74 &148,833 &\$0.32 &92 s \\
HTML &GPT-5 &0.74 &0.83 &314,231 &\$0.62 &624 s \\
HTML &GPT-5-mini &0.57 &0.79 &233,453 &\$0.09 &236 s \\
HTML &Sonnet 4 &0.61 &0.72 &327,640 &\$1.23 &357 s \\
RAG &GPT-4.1 &0.74 &0.86 &\textbf{12,534} &\underline{\$0.03} &\textbf{7 s} \\
RAG &GPT-5 &\underline{0.83} &\textbf{0.96} &100,707 &\$0.18 &123 s \\
RAG &GPT-5-mini &0.78 &\underline{0.93} &32,565 &\textbf{\$0.01} &47 s \\
RAG &Sonnet 4 &\underline{0.83} &0.91 &44,527 &\$0.15 &24 s \\
MCP &GPT-4.1 &0.74 &0.88 &\underline{24,190} &\$0.05 &\underline{10 s} \\
MCP &GPT-5 &\textbf{0.87} &\textbf{0.96} &134,190 &\$0.20 &97 s \\
MCP &GPT-5-mini &0.74 &0.90 &90,948 &\underline{\$0.03} &80 s \\
MCP &Sonnet 4 &0.65 &0.84 &244,762 &\$0.75 &41 s \\
NLWeb &GPT-4.1 &0.57 &0.84 &26,665 &\$0.06 &13 s \\
NLWeb &GPT-5 &\textbf{0.87} &\textbf{0.96} &42,380 &\$0.07 &62 s \\
NLWeb &GPT-5-mini &0.78 &\underline{0.93} &97,861 &\underline{\$0.03} &105 s \\
NLWeb &Sonnet 4 &\underline{0.83} &0.92 &37,005 &\$0.12 &15 s \\
\bottomrule
\end{tabular}
}%
\end{table}

%% file: tables/vague_search.tex
\begin{table}[!htp]\centering
\caption{Results for Vague Product Search. Best bolded, 2nd best underlined.}
\label{tab:complex_search }
\label{tab:vague_product_results}
\resizebox{0.48\textwidth}{!}{%
\begin{tabular}{lrrrrrrr}\toprule
\textbf{Agent} & \textbf{Model} & \textbf{CR} & \textbf{F1} & \textbf{Token} & \textbf{Cost} & \textbf{Runtime} \\\midrule
HTML &GPT-4.1 &0.32 &0.49 &157,639 &\$0.34 &97 s \\
HTML &GPT-5 &0.60 &\underline{0.78} &288,320 &\$0.60 &641 s \\
HTML &GPT-5-mini &0.32 &0.60 &255,852 &\$0.09 &248 s \\
HTML &Sonnet 4 &0.42 &0.60 &337,967 &\$1.25 &361 s\\
RAG &GPT-4.1 &0.26 &0.53 &18,198 &\$0.04 &\textbf{9 s} \\
RAG &GPT-5 &\textbf{0.74} &\textbf{0.82} &111,872 &\$0.20 &147 s \\
RAG &GPT-5-mini &0.58 &0.66 &68,452 &\textbf{\$0.02} &83 s \\
RAG &Sonnet 4 &0.37 &0.41 &92,409 &\$0.30 &42 s\\
MCP &GPT-4.1 &0.11 &0.46 &\textbf{27,456} &\$0.06 &\underline{12 s} \\
MCP &GPT-5 &0.47 &0.65 &154,716 &\$0.23 &119 s \\
MCP &GPT-5-mini &0.53 &0.73 &105,082 &\underline{\$0.03} &101 s \\
MCP &Sonnet 4 &0.32 &0.53 &276,672 &\$0.85 &53 s \\
NLWeb &GPT-4.1 &0.37 &0.60 &\underline{27,689} &\$0.06 &13 s \\
NLWeb &GPT-5 &0.53 &0.72 &77,641 &\$0.12 &62 s \\
NLWeb &GPT-5-mini &\underline{0.63} &0.77 &119,435 &\$0.04 &126 s \\
NLWeb &Sonnet 4 &0.37 &0.55 &62,435 &\$0.20 &26 s \\
\bottomrule
\end{tabular}
}%
\end{table}

%% file: tables/performance-drop-price.tex
\begin{table}[!htp]\centering
\caption{Results for Cheapest Product Search. Best bolded, 2nd best underlined.}
\label{tab:price_drop}
\begin{adjustbox}{width=\columnwidth,center}
\begin{tabular}{lrrrrrrr}\toprule
    \textbf{Agent} & \textbf{Model} & \textbf{CR} & \textbf{F1} & \textbf{Token} & \textbf{Cost} & \textbf{Runtime} \\\midrule
    HTML &GPT-4.1 &0.50 &0.58 &149,657 &\$0.32 &97 s\\
    HTML &GPT-5 &\textbf{0.75} &0.75 &194,594 &\$0.37 &465 s\\
    HTML &GPT-5-mini &0.54 &0.59 &209,629 &\$0.08 &197 s\\
    HTML &Sonnet 4 &0.54 &0.54 &256,543 &\$0.93 &264 s\\
    RAG &GPT-4.1 &0.62 &0.68 &\textbf{16,958} &\$0.04 &\textbf{9 s} \\
    RAG &GPT-5 &\underline{0.72} &\textbf{0.78} &70,736 &\$0.14 &129 s\\
    RAG &GPT-5-mini &0.69 &\underline{0.76} &29,754 &\textbf{\$0.01} &46 s \\
    RAG &Sonnet 4 &0.50 &0.50 &70,887 &\$0.23 &35 s\\
    MCP &GPT-4.1 &0.42 &0.60 &27,816 &\$0.06 &\textbf{9 s} \\
    MCP &GPT-5 &0.69 &0.72 &94,017 &\$0.14 &76 s\\
    MCP &GPT-5-mini &0.54 &0.57 &95,723 &\underline{\$0.03} &75 s\\
    MCP &Sonnet 4 &0.62 &0.63 &217,687 &\$0.67 &41 s\\
    NLWeb &GPT-4.1 &0.27 &0.42 &\underline{25,941} &\$0.05 &\underline{13 s} \\
    NLWeb &GPT-5 &0.65 &0.75 &58,185 &\$0.09 &60 s\\
    NLWeb &GPT-5-mini &0.54 &0.59 &71,877 &\underline{\$0.03} &93 s\\
    NLWeb &Sonnet 4 &0.58 &0.63 &42,790 &\$0.14 &21 s\\
\bottomrule
\end{tabular}
\end{adjustbox}
\end{table}

%% file: tables/trasaction_results.tex
\begin{table}[!htp]\centering
\caption{Results for Action \& Transaction. Best bolded, 2nd best underlined.}
\label{tab:transaction_results}
\resizebox{0.48\textwidth}{!}{%
\begin{tabular}{lrrrrrrr}\toprule
\textbf{Agent} & \textbf{Model} & \textbf{CR} & \textbf{F1} & \textbf{Token} & \textbf{Cost} & \textbf{Runtime} \\\midrule
HTML &GPT-4.1 &\textbf{1.00} &\textbf{1.00} &124,782 &\$0.27 &79 s \\
HTML &GPT-5 &0.67 &0.64 &182,320 &\$0.35 &395 s\\
HTML &GPT-5-mini &0.53 &0.56 &149,165 &\$0.05 &153 s\\
HTML &Sonnet 4 &0.87 &0.87 &196,076 &\$0.71 &189 s\\
RAG &GPT-4.1 &0.87 &0.96 &\textbf{8,445} &\underline{\$0.02} &\textbf{6 s} \\
RAG &GPT-5 &\underline{0.93} &\underline{0.98} &18,631 &\$0.04 &35 s\\
RAG &GPT-5-mini &0.47 &0.54 &\underline{12,380} &\textbf{\$0.01} &26 s\\
RAG &Sonnet 4 &\underline{0.93} &\underline{0.98} &17,635 &\$0.06 &27 s\\
MCP &GPT-4.1 &0.73 &0.86 &45,714 &\$0.09 &\underline{13 s} \\
MCP &GPT-5 &\underline{0.93} &\underline{0.98} &193,062 &\$0.28 &88 s\\
MCP &GPT-5-mini &0.67 &0.86 &289,448 &\$0.08 &160 s\\
MCP &Sonnet 4 &\underline{0.93} &\underline{0.98} &182,728 &\$0.56 &40 s\\
NLWeb &GPT-4.1 &0.87 &0.96 &38,565 &\$0.08 &14 s\\
NLWeb &GPT-5 &\underline{0.93} &\underline{0.98} &83,333 &\$0.14 &50 s\\
NLWeb &GPT-5-mini &0.53 &0.74 &174,717 &\$0.05 &108 s\\
NLWeb &Sonnet 4 &\underline{0.93} &\underline{0.98} &46,833 &\$0.15 &19 s\\
\bottomrule
\end{tabular}
}%
\end{table}

%% file: tables/efficiency-overview.tex
\begin{table}[!htp]\centering
\caption{Efficiency overview by agent and model.}
\label{tab:efficiencyOverview}
\resizebox{0.48\textwidth}{!}{%
\begin{tabular}{lrrrrr}\toprule
    \textbf{Interface} & \textbf{Model} & \textbf{Token} & \textbf{Cost} & \textbf{Runtime} \\\midrule
    HTML &Average &225,090 &\$0.49 &281 s \\
    HTML &GPT-4.1 &146,761 &\$0.32 &92 s \\
    HTML &GPT-5 &253,759 &\$0.50 &522 s \\
    HTML &GPT-5-mini &215,885 &\$0.08 &211 s \\
    HTML &Sonnet 4 &283,956 &\$1.05 &299 s \\
    RAG &Average &47,093 &\$0.10 &51 s \\
    RAG &GPT-4.1 & \textbf{14,477} &\underline{\$0.03} & \textbf{8 s} \\
    RAG &GPT-5 &79,142 &\$0.15 &114 s \\
    RAG &GPT-5-mini &36,252 & \textbf{\$0.01} &51 s \\
    RAG &Sonnet 4 &58,885 &\$0.20 &32 s \\
    MCP &Average &121,624 &\$0.25 &57 s \\
    MCP &GPT-4.1 &29,964 &\$0.06 & \underline{11 s} \\
    MCP &GPT-5 &119,841 &\$0.18 &94 s \\
    MCP &GPT-5-mini &104,319 &\underline{\$0.03} &80 s \\
    MCP &Sonnet 4 &232,374 &\$0.71 &44 s \\
    NLWeb &Average &57,840 &\$0.08 &49 s \\
    NLWeb &GPT-4.1 & \underline{28,876} &\$0.06 &14 s \\
    NLWeb &GPT-5 &56,922 &\$0.09 &59 s \\
    NLWeb &GPT-5-mini &98,449 &\underline{\$0.03} &102 s \\
    NLWeb &Sonnet 4 &46,415 &\$0.15 &20 s \\
\bottomrule
\end{tabular}
}%
\end{table}

%% file: figures/cost-figure.tex
\begin{figure*}[!ht]
    \centering
    \includegraphics[width=0.9\textwidth]{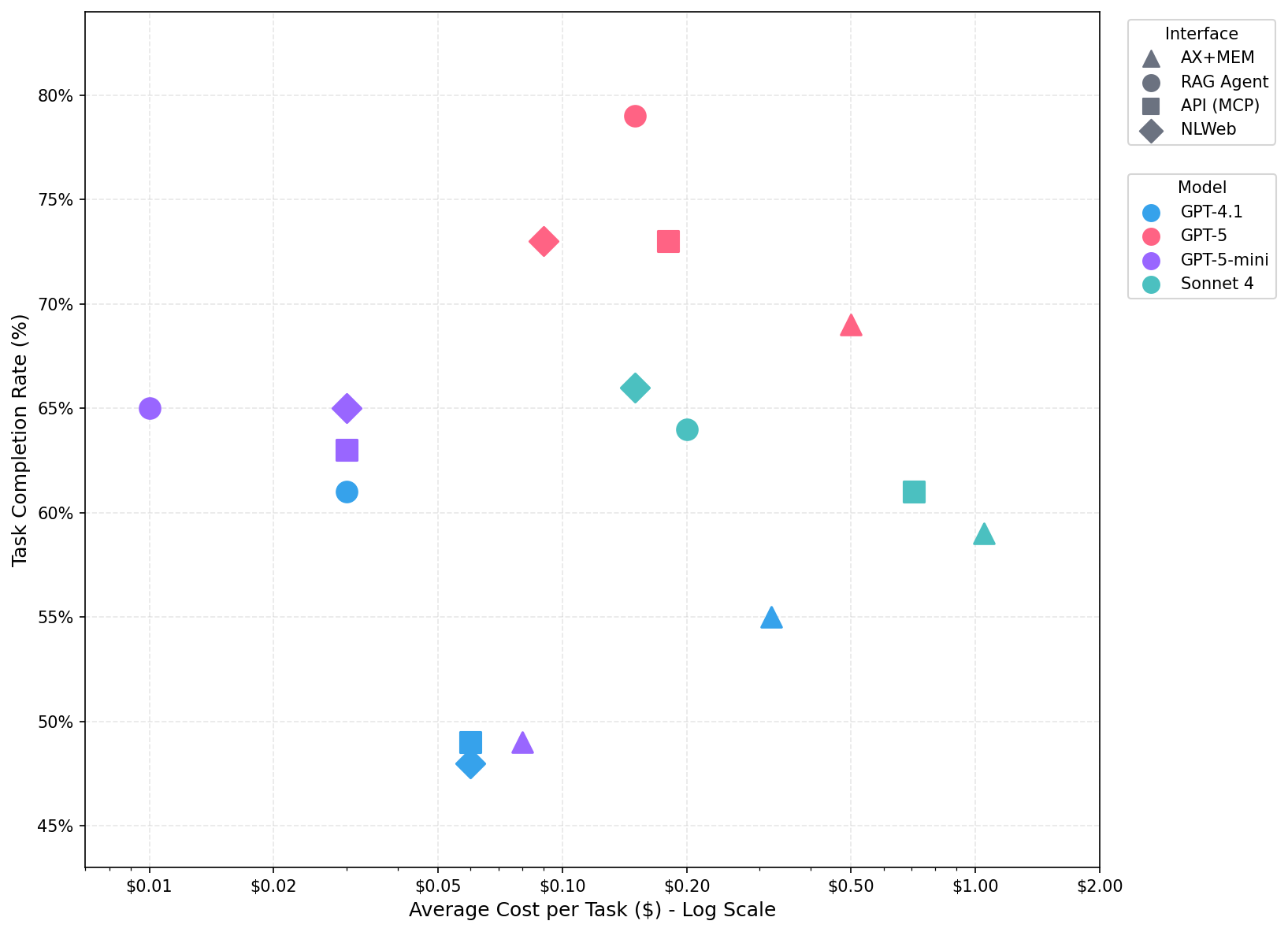}
    \caption{Price (log scale) and performance across architecture and model combinations. Top left is best, meaning higher F1 and lower cost. RAG with GPT 5 mini sits on the cost to quality frontier. RAG with GPT 5 moves right in cost with a smaller gain in F1.}
    \label{fig:price-to-performance-ratio}
\end{figure*}

%% file: sections/05_error_analysis.tex
\section{Error Analysis}
\label{sec:error_analysis}

This section presents an error analysis for the three non-HTML interfaces: RAG, MCP, and NLWeb. The analysis is based on runs with GPT-4.1 and Claude Sonnet 4. Each agent output was compared to the benchmark test set, distinguishing false negatives (FN), where relevant items were missed, and false positives (FP), where irrelevant items were included into the results. False negatives are further divided into non-retrieved cases, where the correct item was never retrieved, and retrieved cases, where it was retrieved but not selected. Each of the 729 false positive errors was manually annotated and grouped into error categories as shown in Table \ref{tab:error-analysis}.

\input{tables/error_categories}

Errors are roughly balanced overall but differ by interface. RAG shows a near even split, with slightly more false negatives. MCP and NLWeb produce more false positives, reflecting reasoning and constraint-handling problems rather than retrieval problems. GPT-4.1 under RAG has the highest share of non-retrieved false negatives, while Claude under NLWeb has the most retrieved false negatives. Claude makes fewer errors it total but misses more correct items, whereas GPT-4.1 adds more incorrect ones.

\textbf{False negatives:} False negatives account for roughly one third of all errors. RAG contributes the largest share, with non-retrieved cases dominating. About half of all non-retrieved errors occur under RAG, indicating a retrieval coverage limitation. This limitation is most pronounced for GPT-4.1 and Claude, which typically query the index once or twice per task. The lower number of retrieved offers per attempt is also reflected in RAG’s reduced token usage. Structured interfaces show the opposite trend. In NLWeb, most false negatives are retrieved, meaning the correct product was retrieved but not recognized. MCP falls between the two, exhibiting an approximately even distribution of retrieved and non-retrieved errors.

\textbf{False positives:} False positives are more frequent overall and vary in type. The largest single class ($\sim$25 \%) is product fails requirements, where items often fit in general but violate a specific requirement such as memory size. Price-related errors are also common, as are variant mismatches such as returning a special edition instead of the standard version of a product. Across interfaces, models show a tendency to accept products as relevant that meet most requirements but differ in a single key attribute. This indicates a lack of strict interpretation of constraints. Manual inspection confirms that these false positives are typically near misses. For example, returning a RAM kit instead of a single stick with equivalent capacity, rather than completely unrelated or misaligned products.

\textbf{Category-specific Differences:} The error distribution  varies systematically by task category:  
\begin{itemize}
    \item In specific product search, RAG is dominated by non-retrieved false negatives (Table~\ref{tab:error-analysis}), while MCP and NLWeb show more retrieved false negatives, where items are retrieved but not selected. Additional responses are frequent under MCP, often involving incorrect variants or missing specifications.
    
    \item In vague product search, overall error rates are higher. NLWeb shows many retrieved false negatives, often coupled with overgeneralization such as offering loosely related items. Subjective misclassifications, for example interpreting color constraints too broadly, occur in both MCP and NLWeb.  

    \item In cheapest product search, price-related errors dominate. MCP and NLWeb often return near-optimal but slightly overpriced offers, and attribute mismatches are frequent, reflecting the difficulty of balancing multiple constraints.  

    \item In transactional tasks, the overall number of errors is low (Table~\ref{tab:error-analysis}, Action \& Transaction rows). Failures rarely involve cart or checkout operations directly but typically result from selecting the wrong product variant.  
\end{itemize}

\textbf{Qualitative Observations:} Two recurring error patterns emerge from manual inspection. First, agents struggle with physical and spatial reasoning: requests for compact keyboards often yielded full-sized options, and shape-based adapter tasks frequently returned visibly dissimilar items. Second, the agents struggled with comparative expressions: In several cases, comparative expressions (e.g., ``more than'', ``less than'') were misinterpreted as equality checks rather than inequality constraints. 

Overall, most errors are nuanced rather than outright failures, often involving near-miss candidates that partially meet task constraints. These findings suggest two areas for improvement:
(i) increasing retrieval coverage, especially in less well-defined scenarios, and
(ii) applying lightweight validation checks, such as price or attribute thresholds, before finalizing selections.

%% file: tables/error_categories.tex
\begin{table*}[htbp] 
\centering
\caption{Error Counts by error category, interface, and model. Highest per interface and model bolded, 2nd underlined.
}
\label{tab:error-analysis}
\footnotesize
\begin{adjustbox}{max width=\textwidth}
\setlength{\tabcolsep}{6pt}%
\renewcommand{\arraystretch}{0.9}%
\setlength{\aboverulesep}{2pt}\setlength{\belowrulesep}{2pt}%
\begin{tabular}{lllcccccc}
\toprule
\multirow{2}{*}{Task Type} & \multirow{2}{*}{Error Type} & \multirow{2}{*}{Error Category} & \multicolumn{2}{c}{RAG} & \multicolumn{2}{c}{MCP} & \multicolumn{2}{c}{NLWeb} \\
\cmidrule(lr){4-5} \cmidrule(lr){6-7} \cmidrule(lr){8-9}
 &  &  & GPT-4.1 & Sonnet 4 & GPT-4.1 & Sonnet 4 & GPT-4.1 & Sonnet 4 \\
\midrule

\multirow{10}{*}{Specific Product Search}
 & False Positive & Wrong product selected & 3 & 5 & 20 & \textbf{22} & 6 & \underline{20} \\ 
 &  & Wrong product category & 0 & 0 & 7 & \textbf{16} & 0 & \underline{7} \\
 &  & Wrong product variant & 0 & 0 & \textbf{14} & 4 & 5 & \underline{5} \\ 
 &  & Missing product info & 6 & 0 & \textbf{23} & 4 & \underline{22} & \underline{22} \\
 &  & Large specification mismatch & 6 & 4 & 8 & 3 & \textbf{12} & \underline{8} \\
 &  & Price too high & 3 & 3 & \textbf{6} & 0 & \underline{3} & 1 \\
 &  & Product fails requirements & 3 & 0 & \textbf{11} & 9 & \underline{9} & 5 \\
 &  & Ambiguous / unclear & 0 & 2 & \textbf{9} & 3 & 6 & \underline{6} \\
\cmidrule(lr){2-9}
 & False Negative & Non-retrieved & \textbf{28} & 17 & 15 & 12 & \underline{17} & 11 \\ 
 &  & Retrieved & 1 & 10 & 2 & \textbf{12} & 5 & \underline{10} \\

\midrule
\multirow{8}{*}{Vague Product Search}
 & False Positive & Wrong product selected & 1 & 0 & \underline{3} & 1 & \underline{3} & \textbf{4} \\ 
 &  & Wrong product variant & 1 & 1 & 0 & \underline{6} & 0 & \textbf{7} \\ 
 &  & Price too high & 1 & 1 & \textbf{10} & 1 & \underline{7} & 0 \\
 &  & Product fails requirements & 12 & 8 & 31 & \textbf{40} & 15 & \underline{31} \\
 &  & Subjectively wrong & 8 & 0 & \underline{10} & \textbf{11} & \textbf{11} & \textbf{11} \\
 &  & Ambiguous / unclear & 5 & 3 & \textbf{10} & 5 & \underline{5} & \underline{5} \\ 
\cmidrule(lr){2-9}
 & False Negative & Non-retrieved & \textbf{27} & \underline{18} & 9 & 9 & 6 & 8 \\
 &  & Retrieved & 3 & \textbf{26} & 8 & 13 & 12 & \underline{14} \\

\midrule
\multirow{6}{*}{Cheapest Product Search}
 & False Positive & Price too high & 2 & 0 & 9 & 1 & \textbf{19} & \underline{9} \\ 
 &  & Slightly too expensive (5\%) & 3 & 0 & \textbf{8} & 0 & 7 & \underline{7} \\ 
 &  & Product fails requirements & 0 & 1 & \textbf{3} & 1 & 2 & \underline{2} \\
 &  & Ambiguous / unclear & 2 & 3 & 4 & 1 & \textbf{7} & \underline{4} \\
\cmidrule(lr){2-9}
 & False Negative & Non-retrieved & 4 & 1 & 2 & 3 & \textbf{5} & \underline{4} \\
 &  & Retrieved & 0 & \textbf{11} & 3 & 3 & \underline{4} & 3 \\

\midrule

 \multirow{4}{*}{Action \& Transaction} 
 & False Positive & Wrong product variant & 3 & 1 & \textbf{5} & 1 & \underline{4} & 1 \\
 &  & Ambiguous / unclear & 1 & 1 & 1 & \textbf{2} & 1 & \textbf{2} \\ 
\cmidrule(lr){2-9}      
 & False Negative & Non-retrieved & 0 & \textbf{1} & 0 & \textbf{1} & 0 & \textbf{1} \\
  & & Retrieved & 0 & 0 & \textbf{6} & 1 & 1 & \underline{3} \\

\midrule

 & \multicolumn{2}{l}{\textbf{Total False Positives}} & 60 & 36 & \textbf{199} & 131 & \underline{144} & 123 \\
 & \multicolumn{2}{l}{\textbf{Total False Negatives}} & \underline{63} & \textbf{84} & 45 & 54 & 48 & 52 \\
\cmidrule(lr){2-9}
 & \multicolumn{2}{l}{\textbf{Overall Total}} & 123 & 120 & \textbf{244} & 185 & \underline{192} & 175 \\
\bottomrule
\end{tabular}
\end{adjustbox}
\end{table*}

%% file: sections/97_relatedwork.tex
\section{Related Work}

\textbf{LLM Agents:}
One of the first large-language-model agent frameworks was ReAct \cite{yaoReAct2022}, which combined reasoning traces with task actions to enable adaptive planning. Reflexion \cite{Shinn2023Reflexion} extended this idea with self-evaluation and feedback to refine later decisions. Together, these approaches demonstrate that LLM agents can integrate reasoning with action, providing the foundation for more advanced systems that operate across different types of web interfaces. 

Song et al. \cite{song-etal-2025-beyond} compare HTML browsing with direct API calling. They show that API agents reach 15\% higher success rate on the WebArena benchmark~\cite{Zhou2023WebArena} than HTML agents, given that good APIs are available. DeepShop \cite{Lyu2025DeepShop} compares browsing-based and RAG-based agents on shopping tasks. They report a ten-point F1 improvement for RAG over browsing, while noting declines for more complex comparison tasks. Their experiments are not executed in a controlled environment, but on the live Web which limits the reproducibility of their results. These works confirm that the interface strongly influences agent success but they remain limited to comparing two architectures each. We expand this line of research by providing the first comparison of four architectures (HTML, RAG, MCP, and NLWeb) using identical tasks and a controlled environment that enables the reproduction of the results.

\textbf{Web Benchmarks:} Various benchmarks have been developed to evaluate web agents in controlled settings. WebShop \cite{yao2022webshop} and ShoppingBench \cite{Wang2025ShoppingBench} simulate single stores with large product catalogs, supporting reproducibility but do not require cross-shop comparisons. WebArena \cite{Zhou2023WebArena} and REAL \cite{gargREALBenchmarkingAutonomous2025} broaden tasks to multiple domains, yet each task needs to be executed against a single website, avoiding comparison-shopping scenarios. Mind2Web \cite{deng2023mind2web} provides over 2,000 tasks across 137 websites, emphasizing generalist web agents but not isolating the effect of different web interfaces. More recently, DeepShop \cite{Lyu2025DeepShop} introduced a shopping benchmark which requires agents to perform complex product searches on the live Web. 

\textbf{FIPA and OWL Agents:} Petrova et al. \cite{petrova2025semanticwebmasagentic} put the current developments around LLM-based web agents into the broader historical context of the FIPA standards and OWL-based web agents.


%% file: sections/98_conclusion.tex
\balance

\section{Conclusion}
\label{sec:conclusion}
We introduce a testbed for comparing different interfaces that agents use to interact with websites. We used this testbed to compare four architectures: HTML browsing, an index-based RAG architecture, and two API-based architectures (MCP and NLWeb). Across models and tasks, the RAG, MCP, and NLWeb agents outperform the HTML agent by 9 F1 points on average. The gap is largest for search tasks with clearly specified requirements. All agents showed weaknesses on challenging tasks, such as vague or cheapest product searches. Efficiency gains are even stronger: The RAG and NLWeb agents require two to five times fewer tokens on search-oriented tasks, resulting in substantially shorter execution times. These improvements mainly stem from reduced input tokens and the removal of browsing overhead.
In summary, our results show that API-based architectures are more effective and more efficient compared to HTML-browsing. However, offering such APIs requires additional development and maintenance effort, making widespread adoption uncertain. For situations where offering APIs is not feasible, crawling web content and accessing it via RAG has proved to be an effective and efficient alternative in our experiments.

